\documentclass[letterpaper, 10 pt, conference]{ieeeconf}  

\IEEEoverridecommandlockouts                              

\usepackage{graphicx} 
\usepackage[fleqn]{amsmath}
\usepackage{support-caption}
\usepackage{subcaption}

\usepackage[hidelinks]{hyperref}

\usepackage{xcolor} 

\title{\LARGE \bf
In-Hand Pose Estimation and Pin Inspection for Insertion of Through-Hole Components
}
\author{Frederik Hagelskjær and Dirk Kraft
\thanks{
All authors are with The Maersk Mc-Kinney Moller Institute, University of Southern Denmark.\newline
{\tt\small \{frhag,kraft\}@mmmi.sdu.dk}
}
}

\begin{document}

\maketitle
\thispagestyle{empty}
\pagestyle{empty}

\begin{abstract}
The insertion of through-hole components is a difficult task. As the tolerances of the holes are very small, minor errors in the insertion will result in failures. These failures can damage components and will require manual intervention for recovery. Errors can occur both from imprecise object grasps and bent pins. Therefore, it is important that a system can accurately determine the object's position and reject components with bent pins.

By utilizing the constraints inherent in the object grasp a method using template matching is able to obtain very precise pose estimates. Methods for pin-checking are also implemented, compared, and a successful method is shown.

The set-up is performed automatically, with two novel contributions. A deep learning segmentation of the pins is performed and the inspection pose is found by simulation. From the inspection pose and the segmented pins, the templates for pose estimation and pin check are then generated.

To train the deep learning method a dataset of segmented through-hole components is created. The network shows a 97.3~\% accuracy on the test set. The pin-segmentation network is also tested on the insertion CAD models and successfully segment the pins.

The complete system is tested on three different objects, and experiments show that the system is able to insert all objects successfully. Both by correcting in-hand grasp errors and rejecting objects with bent pins.
\end{abstract}

\section{Introduction}

\begin{figure}[tb]
    \begin{center}
    \begin{subfigure}[t]{.23\textwidth}
      \centering
      \includegraphics[trim=0 0 0 2,clip,width=0.99\linewidth]{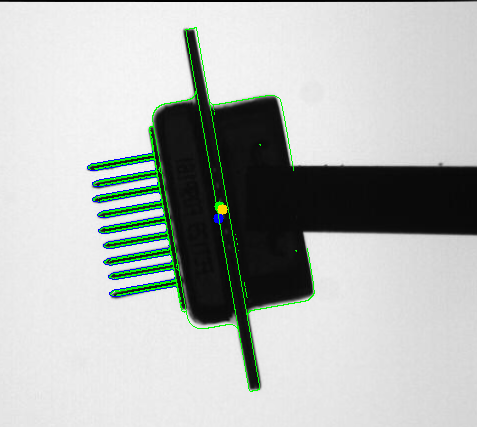}
      \caption{Pose estimation of an object with large angular error compared with the finger.}
      \label{fig:example:1}
    \end{subfigure}%
    ~
    \begin{subfigure}[t]{.23\textwidth}
      \centering
      \includegraphics[trim=0 0 0 4,clip,width=0.99\linewidth]{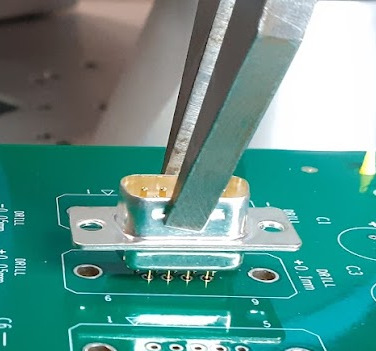}
      \caption{During insertion the finger is rotated to compensate for the large angular error.}
      \label{fig:example:2}
    \end{subfigure}%

    \caption{ Example of the system's ability to estimate the in-hand pose, correct the erroneous grasps, and successfully insert the component.   }
    \label{fig:example}
    \end{center}
\end{figure}

This paper presents a method for post-grasp pose estimation and pre-insertion inspection. The system performs in-hand pose estimation and pin inspection of through-hole components. An example of correcting an erroneous grasp is shown in Fig.~\ref{fig:example}.

The insertion of objects is an essential task in robotics. For insertions to succeed, very precise position information of the object is generally required, and the object should not be damaged.
Several methods have been developed to obtain precise poses of objects. One approach is to focus on restricting the object \cite{mathiesen2018optimisation} or aligning the object with mechanical solutions \cite{wolniakowski2018compensating,hagelskjaer2019combined}. 
The challenge with such methods is that new hardware is needed when new objects are introduced, thus limiting the flexibility of the system.

A different approach is visual pose estimation of the object. This is generally done pre-grasping and has shown very good results in recent years \cite{choi2016using, hodan2020bop}. As these methods are used pre-grasping, any errors introduced by the grasping is not compensated. While using these methods post-grasping is possible, they are challenged by the occlusion presented by the fingers and a general lack of precision needed for very accurate manipulation \cite{hagelskjaer2019using}. 

Methods have been developed specifically for post-grasp pose estimation \cite{choi2016using, wen2020robust}, these methods use depth information that is difficult to obtain with small objects and thus these methods typically focus on larger objects. 
This paper introduces a method using 2D data for in-hand pose estimation of very small objects. The method uses stable poses introduced implicitly by the grasping to return very precise pose estimations.

For the manipulation to be successful, the objects should also be in satisfactory condition. If parts are damaged and the object does not correspond to the rigid CAD model, the manipulation can fail.

The focus of this paper is the insertion of through-hole components. As the tolerances of the holes are tiny, the precision of the in-hand pose must be very high, and the pins cannot be bent.

Several methods have been developed for pin inspection \cite{kashitani1993solder,huang2011application, edinbarough2005vision}, however, these methods focus on post-insertion inspection. 
Pre-insertion inspection is difficult because of the unknown object position and unknown background. Our method uses grasping to bring the object to a position with a known background, and the pose estimation gives the precise position of the pins. The system can then inspect each pin individually, and is able to reject components with bent pins successfully.

Another important aspect of a pose estimation systems is the set-up time. The set-up of a pose estimation system is often a manual task, which is very time-consuming \cite{hagelskjaer2017does}. 

Our system introduces a completely automatic set-up procedure based on the CAD model and grasp pose. Deep learning is used to segment pins, which are used to find the best configuration to inspect the pins. Based on this configuration, templates for pose estimation are generated. This procedure gives a set-up time for a new object as less than one minute.

This paper presents the following main contributions.
\begin{itemize}
    \item In-hand pose estimation in 2D using stable grasp pose.
    \item An inspection method to verify pins are straight.
    \item Deep learning segmentation of object pins.
    \item Automatic set-up based on CAD model and grasp pose.
    \item A system for in-hand pose estimation and pin inspection.
\end{itemize}

The remaining paper is structured as follows: We first review related papers in Sec.~\ref{sec:related}. In Sec.~\ref{sec:method}, all parts of the developed method are explained. In Sec.~\ref{sec:evalution}, each part is tested individually, and a final insertion system test is performed. Finally, in Sec.~\ref{sec:conclusion}, a conclusion is given to this paper, and further work is discussed.

\section{Related Work}
\label{sec:related}

This paper presents a novel system for in-hand pose estimation and pin inspection. However, existing methods also approach the same challenges posed by this application. 

A realm of different methods have been developed to obtain precise pose estimation, and many methods have also been presented for pin inspection. 
In the following, these methods will be detailed, and the reasons that these methods do not solve the specific challenges addressed by the proposed system are discussed.


\subsection{Pose Estimation}

One method to obtain accurate pose estimation is mechanical fixtures. Here very precise fixtures are created, and the objects are placed manually to ensure correct positions. The problem with this solution is that fixtures must be created for each object used, meaning the fixtures must either be refilled often or take up large amounts of space. Additionally, manual labour is needed to fill the fixtures. 

A more autonomous solution is mechanical feeders \cite{mathiesen2018optimisation}. Here, a large number of objects are inserted and returned in the correct pose by mechanical design. 
A different approach to ensure that the object pose is known is to create fingers that compensate for errors \cite{wolniakowski2018compensating}. The geometry of these fingers guides the object into a fixed position, thus removing minor errors. 
This approach can be combined with the vision to optimize the fingers to specifically optimize for the vision errors \cite{hagelskjaer2019combined}. 
The disadvantage of these mechanical solutions is that new hardware must be created for new objects. This severely limits the flexibility of the systems. 


Visual solutions also exist to obtain precise poses. In visual servoing \cite{yu2019siamese}, the robot is continuously updating the position so that the grasped object is in the desired pose. This is usually performed during the insertion. However, this does not allow us to perform the pin inspection. Such methods also rely on training data for the set-up, increasing the set-up time. 
The object pose can also be obtained by tracking the object before and after grasping \cite{liang2020hand}. 
However, for this approach to work, the object must continuously be observed, which is hindered by the in-hand occlusion, and the obtained precision is not high enough for our application.

Another method to predict the in-hand pose is using physical contact \cite{von2020contact}. Here the hand is moved towards the table until the object collides. The position of the collision is used to determine the pose using a particle filter. This approach has also been extended with a visual check \cite{von2021precise}. However, this approach is not feasible for our system as the collisions with the table can bend the pins.

Poses can also be obtained directly with visual pose estimation. 
Methods for 6D pose estimation have shown very good results in recent works \cite{hagelskjaer2022parapose, hodan2020bop}. However, these methods have challenges with in-hand pose estimation as the occlusion by the fingers are not accounted for in the method. Additionally, these methods often lack the very high precision needed for the insertion in our use case \cite{hagelskjaer2019using}.

To solve this, methods have been developed specifically for in-hand pose estimation \cite{choi2016using} \cite{wen2020robust}. Similar to our method, the hand is removed from the detection. However, these methods handle larger objects than those in our use case. They are based on depth information for the pose estimation, which is difficult to obtain for very small objects. 
%
%

An approach similar to ours place the object on a table, this table is then used as a constraint 
\cite{hagelskjaer2019using}. Like our method, search is restricted to 3D as constraints related to the object's movement are used. However, dissimilar to our method, the used constraint is the position on the table, where our method uses the constraint of the grasp. This gives our system greater flexibility as the object can be placed in front of the camera for pin inspection. 
%
%
The pose estimation developed for this paper is based on a previous method \cite{ulrich2012combining}. Here templates are generated based on the object's CAD model and matched in the image using gradients \cite{steger2002occlusion}. However, this method performs 6D pose estimation without constraints, giving uncertainty in the pose estimation, especially in the depth dimension \cite{drost2017introducing}. Our method can obtain a higher precision as movement is allowed in fewer dimensions.

\subsection{Pin Inspection}

While in-hand pin inspection is a novel approach, pin inspection is a classic problem. Inspection is generally performed post-insertion on Printed Circuit Boards (PCB) \cite{kashitani1993solder}. Several methods have been developed with cameras placed above the board \cite{becker2009optical}. One approach is to extract edges using the Canny Edge detector \cite{canny1986computational}, and then measure the position \cite{huang2011application}. This measurement is similar to one of the approaches we test in this paper. However, we also test the match with edge gradients \cite{steger2002occlusion}.
A different approach to ours is pin inspection using neural networks \cite{edinbarough2005vision}. Here a object specific network is trained to classify if pins are bent or not. 
We do not train object specific neural networks in our approach, as training data is not available. 
As opposed to these methods our inspection is performed pre-insertion.

\begin{figure}[bht]
    \vspace{1.5mm}
    \begin{center}
    \includegraphics[width=1.0\linewidth]{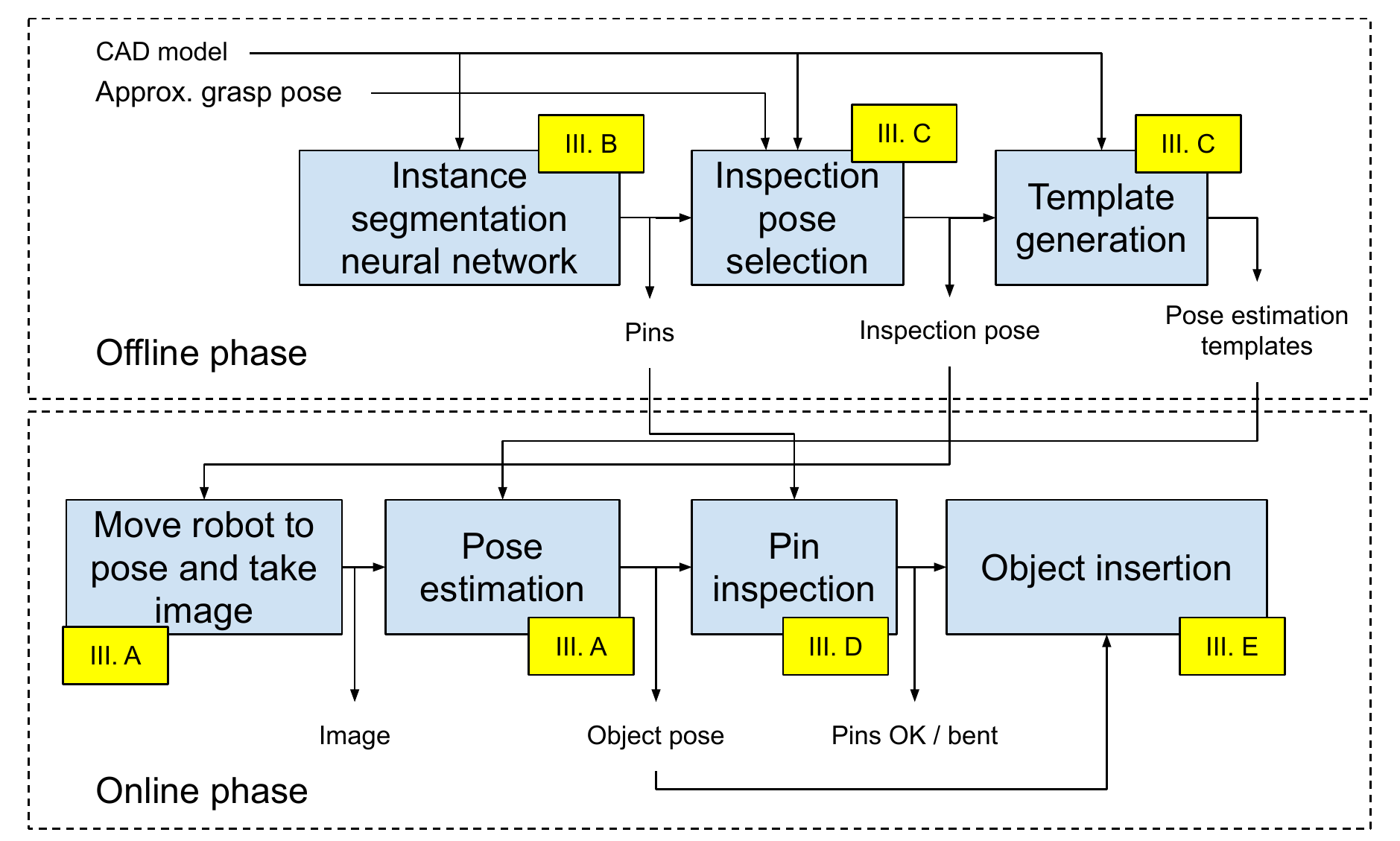} 
    \caption{Overview of the developed system. Yellow boxes link to the chapter discussing the respective component in more detail.}
    \label{fig:system-overview}
    \end{center}
\end{figure}

\section{Method}
\label{sec:method}

The developed system is able to correct for uncertainties in a grasp and inspect if pins are straight for insertion. The full pipeline consists of two different parts. An offline phase for set-up of the system and an online pose estimation and pin inspection phase. Fig.~\ref{fig:system-overview} gives an overview of the steps and the dataflow in the two phases.

The off-line phase requires a CAD model and an approximate grasp pose relative to the object's CAD model. The first step is instance segmentation of the pins. This is performed with a neural network trained on a novel dataset created for pin segmentation. Based on the pin segmentation and grasp poses different inspection poses are checked and the best pose is used to generate templates for the pose estimation.

In the online phase, a pose estimation is first performed to find the object. Using the pose estimation each individual pin is inspected. If all pins are determined to be straight the object can be inserted based on the pose estimation.

\subsection{Pose Estimation}

\begin{figure}[tb]
    \vspace{1.5mm}
    \begin{center}
    \includegraphics[width=0.9\linewidth]{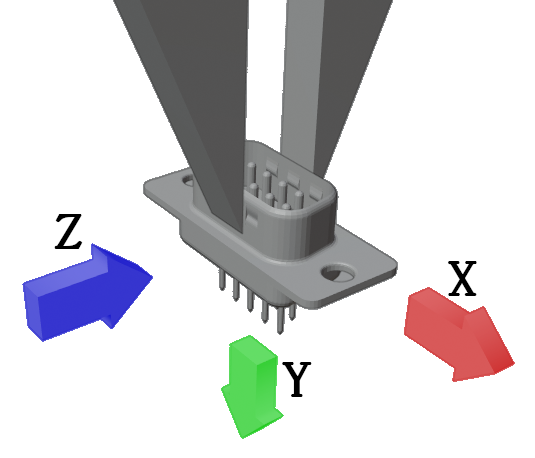} 
    \caption{ Visualization of the stability of the in-hand object grasp. The stable pose of the grasp limits the transform of the object, to translation in X and Y, and rotation in Z, shown respectively by red, green and blue arrows. }
    \label{fig:system}
    \end{center}
\end{figure}

Pose estimation is performed to correct for offsets introduced while grasping the object. To perform the pose estimation, an image is captured where the robot has moved the TCP in front of the camera. 
The pose estimation used in this work is based on an existing method for 6D pose estimation \cite{ulrich2012combining}. 
This method is a template matching approach, based on object CAD model. Edge gradients are matched in the image, and a pyramid search is performed to reduce the run-time. Finally, pose optimization is performed to refine the translation.

However, 6D pose estimation with 2D template matching has shown errors as the depth is unknown \cite{drost2017introducing}. 
Using stable poses such as table constraints has been shown to improve precision drastically for 2D template matching \cite{hagelskjaer2019using}. 

While a table constraint cannot be used in this application, the finger grasp itself creates a stable pose. The geometry of the fingers and the objects align the object in the fingers. A planar grasp surface is created which is oriented in the Z-direction and as such restricts the translation and rotation of the object. As shown in Fig.~\ref{fig:system}, the in-hand object is restricted to translation in X and Y and rotation in the Z-direction. 
The templates for matching are created using these constraints, and this allows for a much smaller search space. This smaller search space improves the precision as wrong detections cannot be made in the restricted directions. The run-time is also decreased as fewer templates are searched.

Additionally, as the position of the fingers grasping the object is known, this "image area" can be excluded from the template model to increase the precision. 

\subsection{Pin Segmentation}
\label{method:segmentation}
For the full method to work, each pin should be inspected individually. 
%
An instance segmentation of the pins is therefore necessary.
While this could be performed manually, this is a time-consuming and challenging process for an untrained user, and errors could be introduced by mislabeling pins. The segmentation of pins is therefore performed automatically. 
The process is split into a segmentation using deep learning, and then an instance segmentation performed with the geometry.

\subsubsection{Network Structure}
The object's CAD model is converted to a point cloud, as several methods for processing 3D data have shown promising performance on point clouds \cite{guo2020deep}.
To include details of the small pins, very dense and thereby large point clouds of the object are needed. 
The pin processing is, therefore, performed using a network that has shown good performance for very large point clouds \cite{hagelskjaer2022deep}. 
The network has also shown an ability to focus on small details, which fits well with the small pins in our application.
As the network in \cite{hagelskjaer2022deep} was created for classification, the segmentation layer from  PointNet \cite{qi2017pointnet} is added to enable instance segmentation.
The training parameters are identical to the original paper \cite{hagelskjaer2022deep}, without the use of error weighing.

\subsubsection{Training Data}
A dataset of components with labelled pins has been created to train the network. The components are downloaded from the free objects from pcb-3d\footnote{\url{https://www.pcb-3d.com/membership_type/free/}}, using only objects labelled "through-hole". The pin labelling was performed manually, with each pin segmented. The dataset consists of 895 objects, with an 85/15 training/test split. With 21~\% of the points labelled pin. The pin segments and code for generating the data are available online at \footnote{\url{https://github.com/fhagelskjaer/pin-segmentation}}.

\subsubsection{Processing CAD model}
\begin{figure}[tb]
    \begin{center}
    \begin{subfigure}[t]{.20\textwidth}
      \centering
      \includegraphics[trim=0 0 0 0,clip,width=0.99\linewidth]{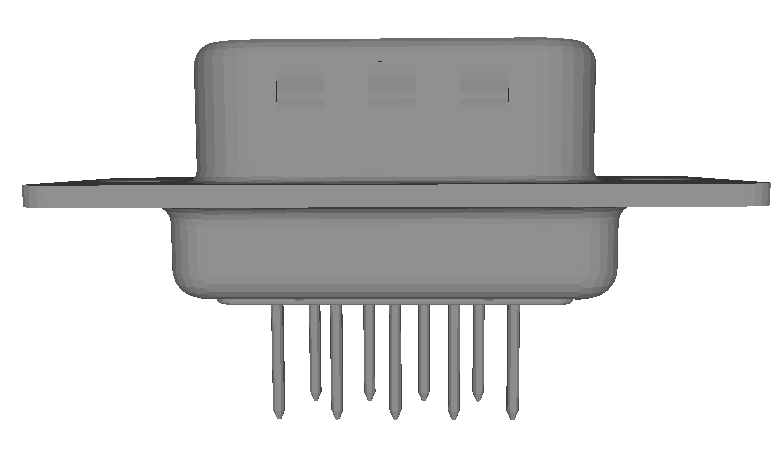}
      \caption{CAD model.}
      \label{fig:segmentation:1}
    \end{subfigure}%
    ~
    \begin{subfigure}[t]{.20\textwidth}
      \centering
      \includegraphics[trim=600 250 400 70,clip,width=0.99\linewidth]{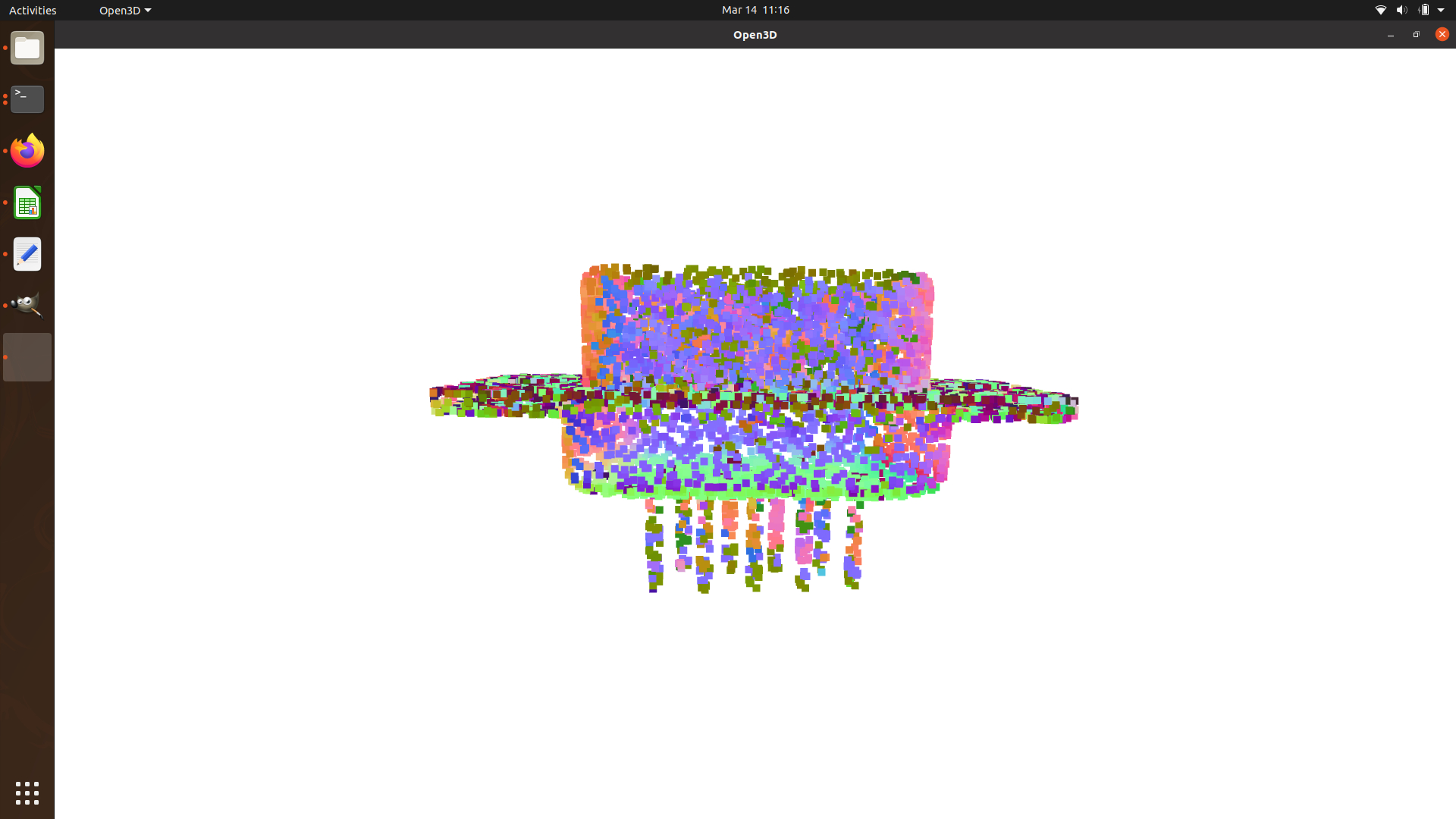}
      \caption{Point Cloud Representation.}
      \label{fig:segmentation:2}
    \end{subfigure}%

    \begin{subfigure}[t]{.20\textwidth}
      \centering
      \includegraphics[trim=600 250 400 70,clip,width=0.99\linewidth]{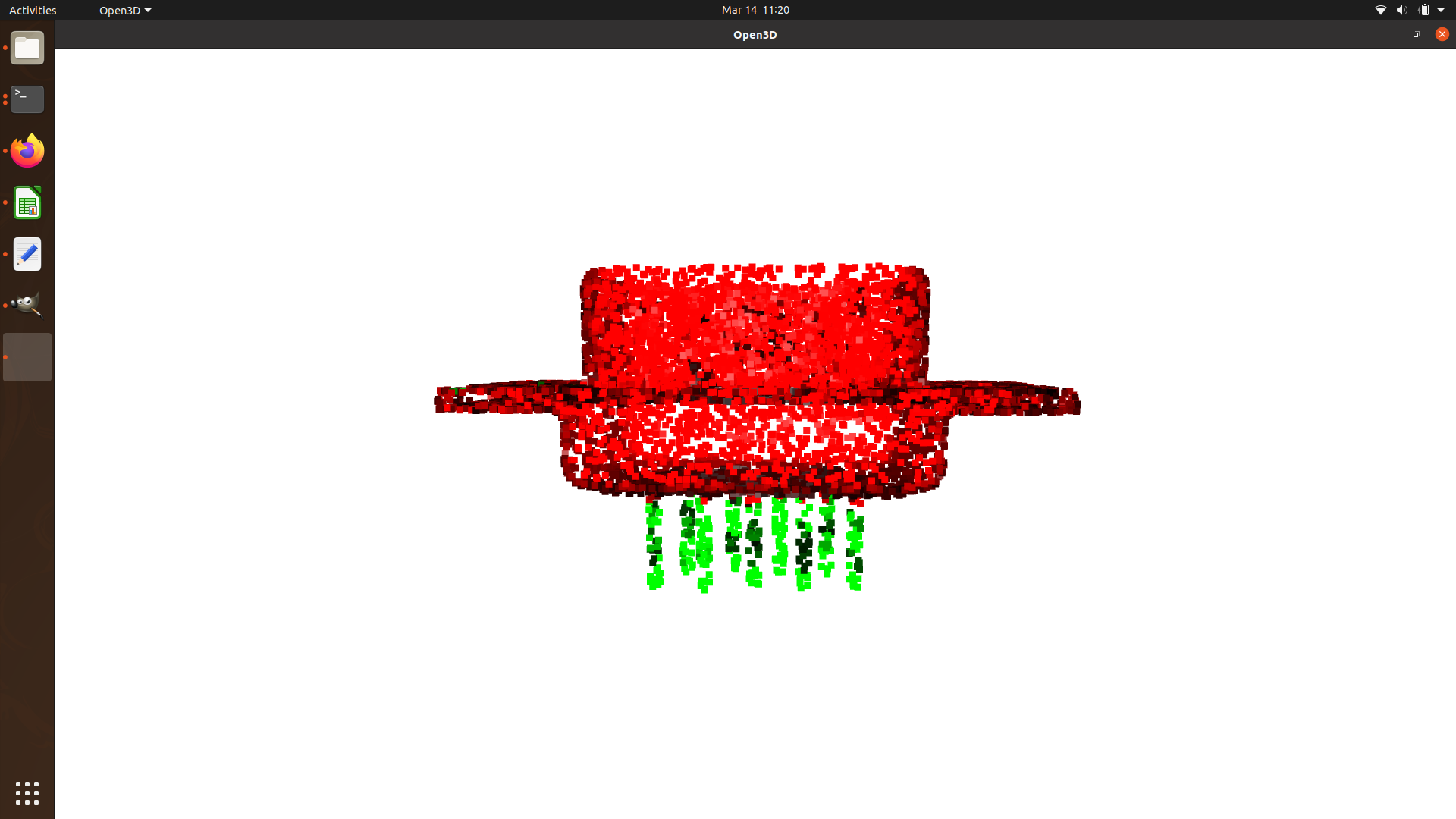}
      \caption{Pin Segmentation.}
      \label{fig:segmentation:3}
    \end{subfigure}%
    ~
    \begin{subfigure}[t]{.20\textwidth}
      \centering
      \includegraphics[trim=50 85 50 50,clip,width=0.99\linewidth]{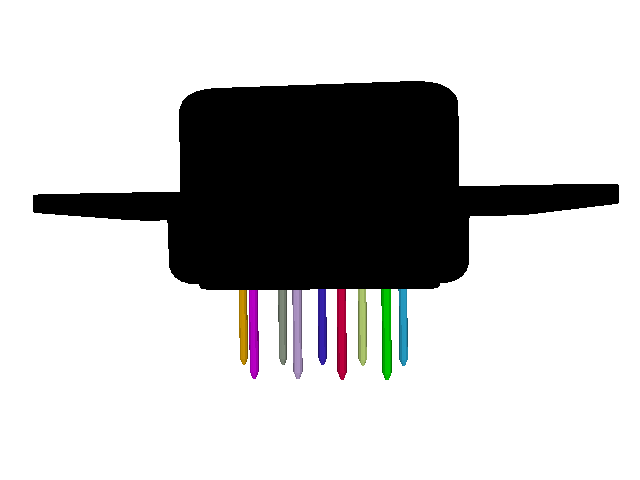}
      \caption{Instance Segmentation.}
      \label{fig:segmentation:4}
    \end{subfigure}%
    
    \caption{ Point cloud segmentation and final instance segmentation of pins. }
    \label{fig:segmentation}
    \end{center}
\end{figure}

When processing a CAD model it is first converted to a point cloud and segmentations are computed using the neural network.
The segmentations in the point cloud are then fit to the original CAD model. 
For each vertex in the CAD model a kNN with $k=15$ is used to find the nearest points in the point cloud. Fifteen points are chosen to smooth the prediction and compensate for errors.  
On the CAD model, connected components are then used to compute instance segmentations of each pin. To ensure that only pins used for insertion are inspected, pins should be in the positive Z-direction of the TCP. The resulting segmentation procedure for object \textbf{D-Sub} is shown in Fig.~\ref{fig:segmentation}. 



\subsection{Generating Pose Checks}

The object is moved in front of the camera to perform the pose estimation and pin inspection. The object's position in front of the camera is known, as the robot is calibrated to the camera, and the grasp pose is known. 

For the pin inspection to be successful, all individual pins should be visible for the camera. To select the best inspection pose, an occlusion test of all pins is performed. This test is performed in simulation with the instance segmentation obtained in Sec.~\ref{method:segmentation}. The object is visualized, and the visibility is measured for each pin. The visibility is measured by first projecting the individual pin into the scene and then projecting the entire object and measuring the occlusion of the pin. To be able to use background segmentation for the pin inspection, the pin should also not occlude the object. 

\begin{figure}[tb]
    \vspace{1.5mm}
    \begin{center}
    \begin{subfigure}[t]{.17\textwidth}
      \centering
      \includegraphics[trim=150 20 100 50,clip,width=0.99\linewidth]{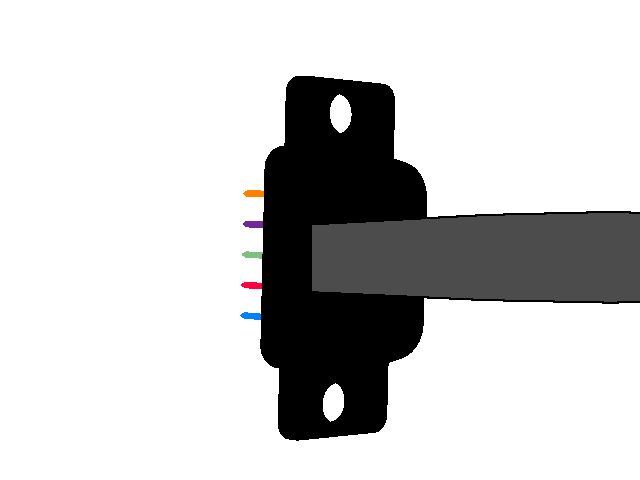}
      \caption{Occluded pins.}
      \label{fig:posecheck:1}
    \end{subfigure}%
    ~
    \begin{subfigure}[t]{.17\textwidth}
      \centering
      \includegraphics[trim=150 20 100 50,clip,width=0.99\linewidth]{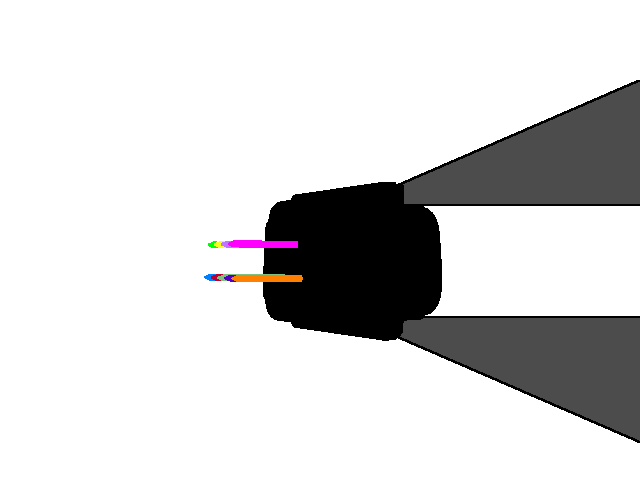}
      \caption{Pins occluding other pins.}
      \label{fig:posecheck:2}
    \end{subfigure}%

    \begin{subfigure}[t]{.17\textwidth}
      \centering
      \includegraphics[trim=150 20 100 50,clip,width=0.99\linewidth]{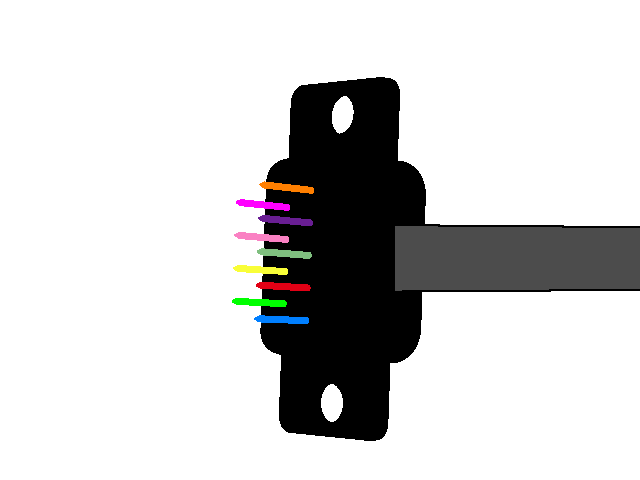}
      \caption{Pins occluding object.}
      \label{fig:posecheck:3}
    \end{subfigure}%
    ~
    \begin{subfigure}[t]{.17\textwidth}
      \centering
      \includegraphics[trim=150 20 100 50,clip,width=0.99\linewidth]{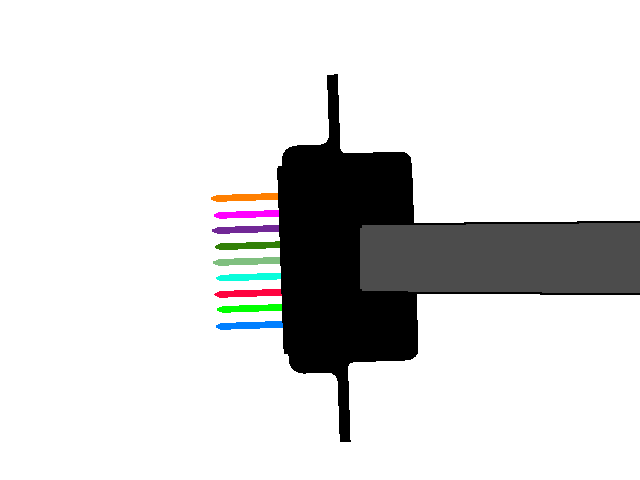}
      \caption{No pin occlusion.}
      \label{fig:posecheck:4}
    \end{subfigure}%
    
    \caption{ Pin occlusion test of pre-configured inspection poses.
    }
    \label{fig:posecheck}
    \end{center}
\end{figure}

As the object can move in-hand, the visibility is checked while rotating the object from -20 to 20 degrees. The minimum visibility when rotating the object is then used.
Six pre-configured inspection poses are tested this way, and the pose with the least occlusion is used. Four different inspection poses are visualized for the \textbf{D-Sub} object in Fig.~\ref{fig:posecheck}.

    

\subsection{Pin Inspection}

To ensure that the object can be inserted, each individual pin needs to be straight. The pin inspection is based on the pose estimation and the instance segmentation. Using the pose estimation each pin is inspected in the image, and the score of the most bent pin is returned. If this score is too low the component cannot be inserted, and a new component is grasped.

Four different pin inspection methods have been developed and are tested on a dataset to find the best method. The four methods are \textit{gradient check}, \textit{intensity overlay}, \textit{edge overlay}, and \textit{distance to edge}. All methods return a score and using the test dataset, a cutoff value is found. 

For the \textit{gradient check}, the same check is performed as in the pose estimation \cite{ulrich2012combining}. In the \textit{intensity overlay} test, a background segmentation is performed, and the amount of pin overlaying the background is measured. For the \textit{edge overlay} edges are computed in the image and the pin edges overlap is measured. For the \textit{distance to edge} the image edges are computed and the distance to pin edges are found. The percentage of edges with less than 2 pixels distance is then returned.

\subsection{Insertion of objects}

The final goal of the system is the insertion of components. To insert the object the \emph{inserted object} position is first found. By inserting the object into the board, and grasping it with the robot the \emph{inserted object} pose $\ _{base}T^{tcp-ins}$ is found. By then moving the robot in front of the camera and obtaining $ \ _{cam}T^{obj}$, the pose $\ _{tcp}T^{obj}$ is found as in Eq.~\ref{eqn:tcp2obj}.

\begin{equation}
\label{eqn:tcp2obj}
        \ _{tcp}T^{obj} = (\ _{cam} T ^{tcp})^{-1} \ _{cam}T^{obj}
\end{equation}

From this, the position of the \emph{inserted object} in the base frame can be computed.

\begin{equation}
\label{eqn:base2obj}
        \ _{base}T^{obj-ins} = \ _{base} T ^{ins} \ _{tcp}T^{obj}
\end{equation}

When an object is grasped and an insertion should be performed the object is moved in front of the camera and pose estimation is performed. Using the pose estimation the pins are inspected and objects with bent pins are rejected. If the object is accepted, the in-hand position, $\ _{tcp}T^{obj-current}$, is found according to Eq.~\ref{eqn:tcp2obj}. And the new insertion pose can then be calculated as in Eq.~\ref{eqn:compensated}.

\begin{multline}
\label{eqn:compensated}
    \ _{base}T^{tcp-ins-comp} = \\
       \qquad \qquad \ _{base}T^{obj-ins} (\ _{tcp}T^{obj-current})^{-1} \\
\end{multline}


\section{Experiments}
\label{sec:evalution}

To verify the validity of the system, several tests are performed. For each system component, a test has been performed to measure the individual validity. Furthermore, a final system test which grasps the object, perform pose estimation and pin inspection and then inserts the object is reported on.

\subsection{Pin Segmentation}

To process the pin both position and normal-vector information is used for each point in the point cloud. The point clouds are sampled at a size of 8192 points, and the position information is normalized. The pin segmentation network is trained for 200 epochs, with a learning rate of 0.001. Two per cent Gaussian noise is attached to both position and normal vector. 

The final model has an accuracy of 96.5~\% on the training data with augmentation, and a test accuracy without augmentation at 97.3~\%. The individual precision and recall for both object and pin is shown in Tab.~\ref{tab:segmentation}. It is seen that the model obtains a precision of 90~\% and 94.8~\% recall.  
In the training dataset pins represent 17.8~\% of the points. 
To compensate for this, loss weighing \cite{wang2017learning} could be used, but the results are adequate for the use with k-NN.

\begin{figure}[tb]
    \begin{center}
    \includegraphics[trim=600 90 400 70,clip,width=0.45\linewidth]{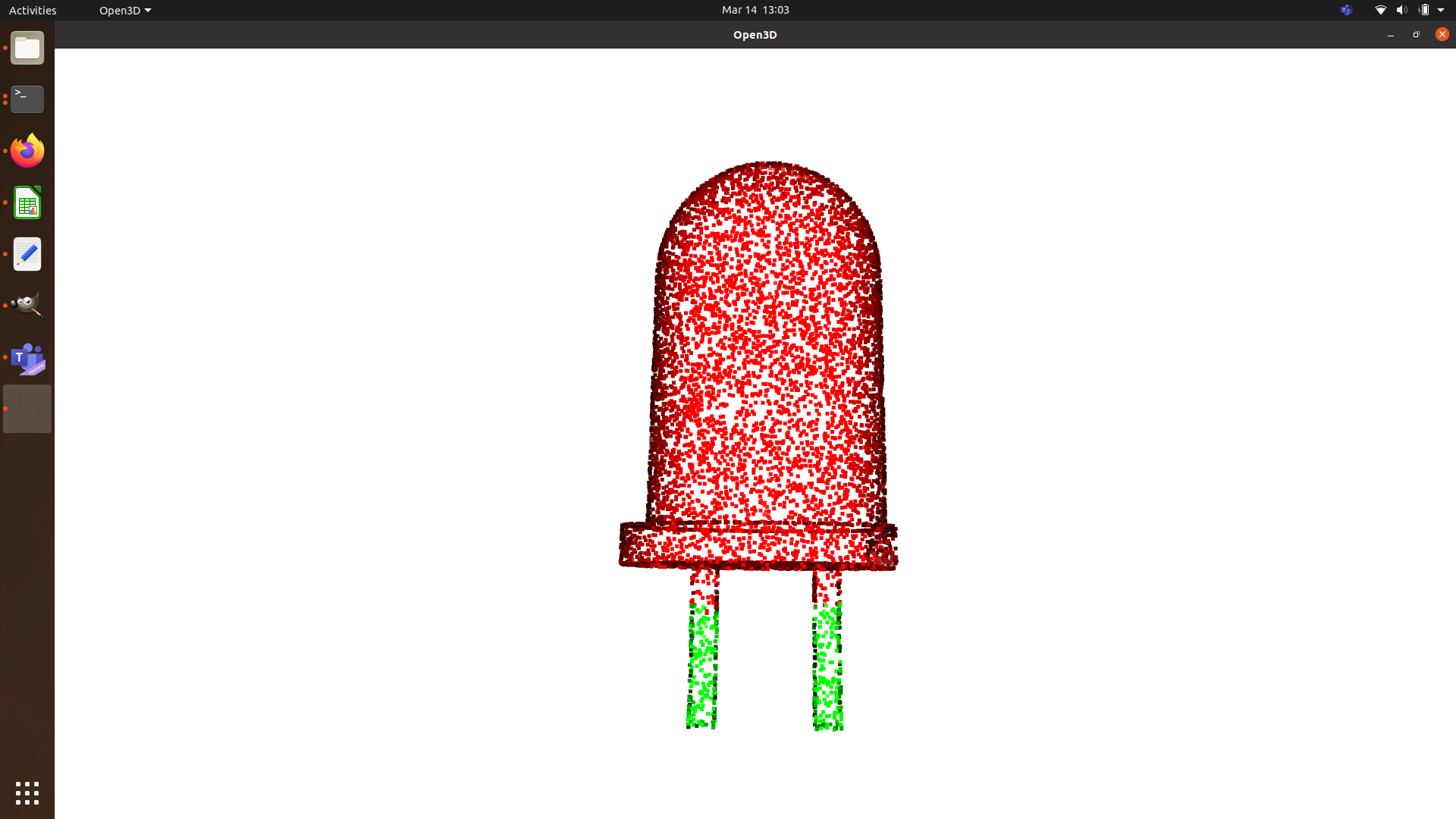} 
    \includegraphics[trim=150 90 150 50,clip,width=0.45\linewidth]{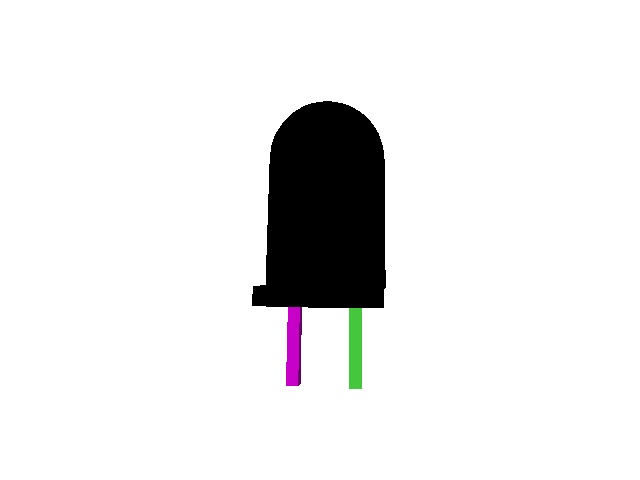} 

    \includegraphics[trim=600 250 400 200,clip,width=0.45\linewidth]{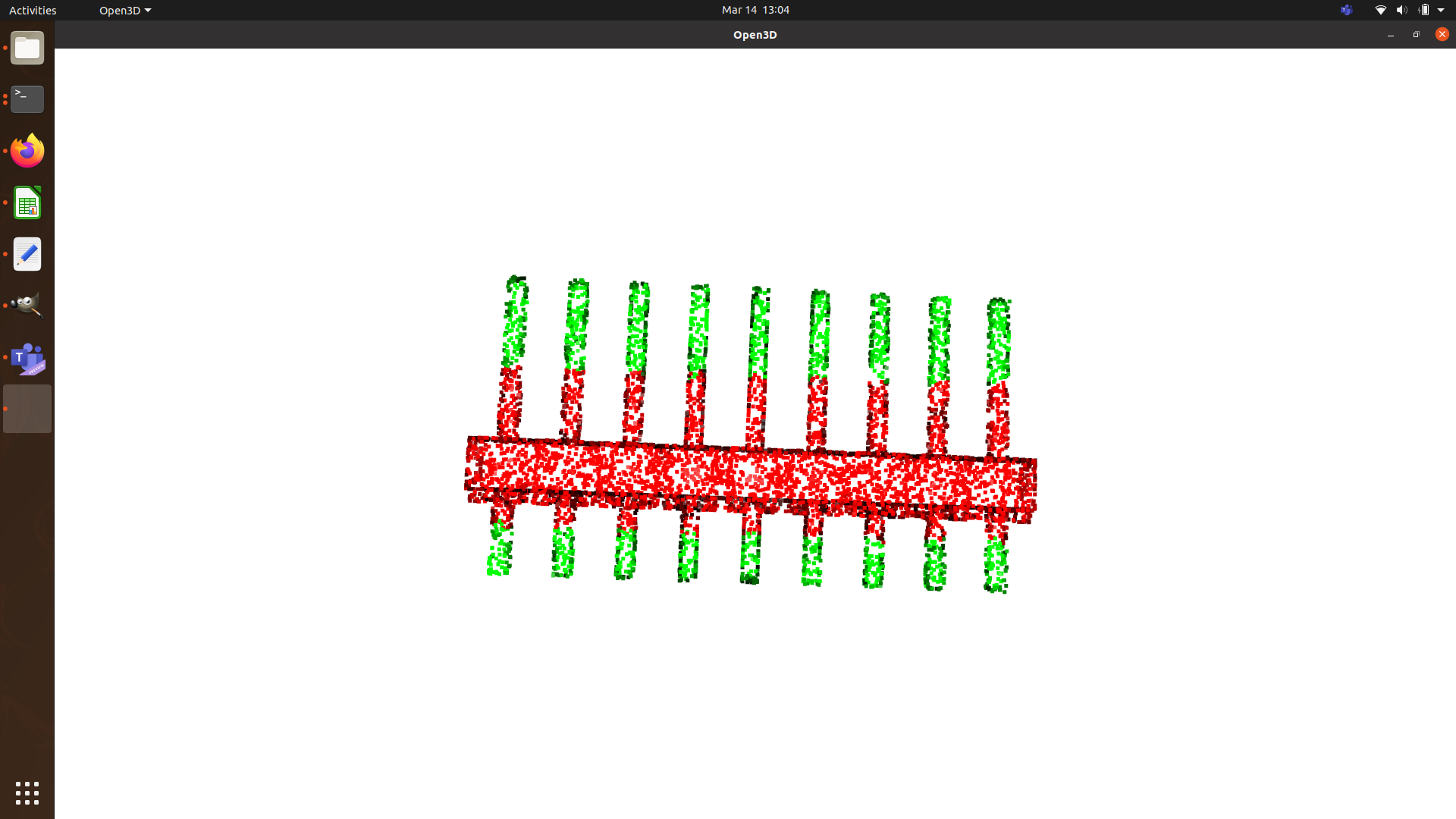} 
    \includegraphics[trim=150 135 150 100,clip,width=0.45\linewidth]{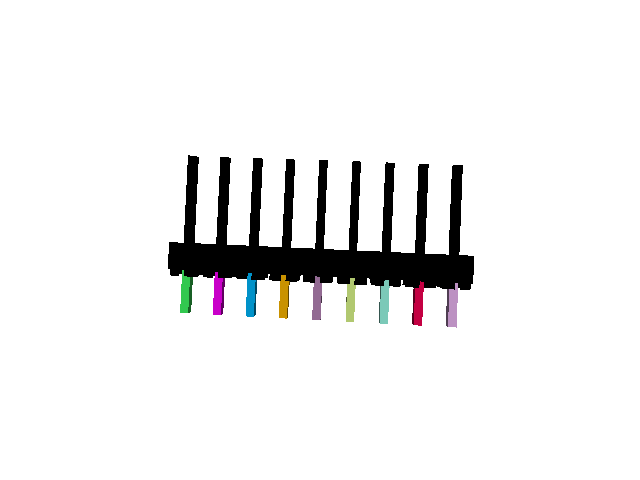} 
    
    \includegraphics[trim=600 250 400 130,clip,width=0.45\linewidth]{gfx/d-sub-seg.png} 
    \includegraphics[trim=50 85 50 50,clip,width=0.45\linewidth]{gfx/segmentation.png} 
    \caption{ Point cloud segmentation and final instance segmentation of pins for the three objects. }
    \label{fig:seg_all}
    \end{center}
\end{figure}

By using k-NN the small amount of point classification errors are removed when segmenting the CAD model. The test objects are all segmented correctly. Segments for the three test objects are shown in Fig.~\ref{fig:seg_all}.

\begin{table}[tb]
    \vspace{1.5mm}
\begin{center}
    \caption{ Precision-Recall for pin segmentation on the test data. }
    \label{tab:segmentation}
    \small
\begin{tabular}{|l|c|c|}
    \hline
         & Precision & Recall \\ \hline
        \textbf{Component} & 98.9 & 97.8 \\ \hline
        \textbf{Pin} & 90.0 & 94.8 \\ \hline
    \end{tabular}
\end{center}
\vspace{-0mm}
\end{table}

\subsection{Pose Generation}

The pose generation step has been introduced to avoid manual set-up, and thereby have a set-up time that enables fast change-overs. When a new CAD model is introduced the system should quickly adapt to the object. All tests, reported on in this subsection, are performed using a Laptop Environment with an Intel Core i7-10610U CPU, with no GPU.

First, the segmentation is performed using the neural network. The pose checks are then performed and finally, the templates for the pose estimation are generated. The run-time of each part for the three objects is shown in Tab.~\ref{tab:setup}. 

For our test objects, the set-up time for a new object is less than one minute. The set-up is, therefore, feasible for fast changeovers in industry. The largest amount of time is spent on the six inspection poses, thus, depending on the set-up time requirements, the number of inspection poses should be limited.

\begin{table}[tb]
\begin{center}
    \caption{ Set-up time for each object. }
    \label{tab:setup}
    \small
\begin{tabular}{|l|c|c|c|}
    \hline
        Object & Segmentation & Pose Check & Template gen. \\ \hline
        \textbf{LED} & 6,059s & 11,572s & 2.949s \\ \hline
        \textbf{Pinheader} & 6,005s & 29,856s & 7.342s \\ \hline
        \textbf{D-Sub} & 6,430s & 40,280s & 7.608s \\ \hline
    \end{tabular}
\end{center}
\vspace{-0mm}
\end{table}

\subsection{Pose Estimation}

To test the precision of the pose estimation a dataset has been created. The dataset consist of each object moved to twelve known offset positions along with a calibration pose. The robot is moved to the calibration pose, $\ _{cam} T ^{tcp}$, and a pose estimate, $_{cam} T ^{obj}$, is found. With these Eq.~\ref{eqn:tcp2obj-calib} is used to find the in-hand position of the object.

\begin{equation}
\label{eqn:tcp2obj-calib}
        \ _{tcp} T ^{obj-calib} = (\ _{cam} T ^{tcp})^{-1} \ _{cam} T ^{obj}
\end{equation}

At each new position, the $\ _{tcp} T ^{obj-calib}$ transform is used to find the expected position in the camera as according to Eq.~\ref{eqn:cam2obj-expected}.

\begin{equation}
\label{eqn:cam2obj-expected}
        \ _{cam} T ^{obj-expected} =  \ _{cam} T ^{tcp-test} \ _{tcp} T ^{obj-calib}
\end{equation}

\begin{figure}[tb]
    \vspace{1.5mm}
    \begin{center}
    \begin{subfigure}[t]{.20\textwidth}
      \centering
      \includegraphics[trim=50 0 50 0,clip,width=0.99\linewidth]{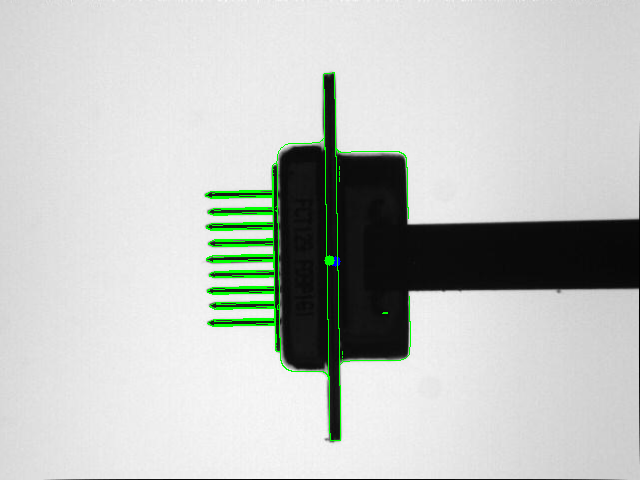}
      \caption{Calibration pose.}
      \label{fig:poseestimation:1}
    \end{subfigure}%
    ~
    \begin{subfigure}[t]{.20\textwidth}
      \centering
      \includegraphics[trim=50 0 50 0,clip,width=0.99\linewidth]{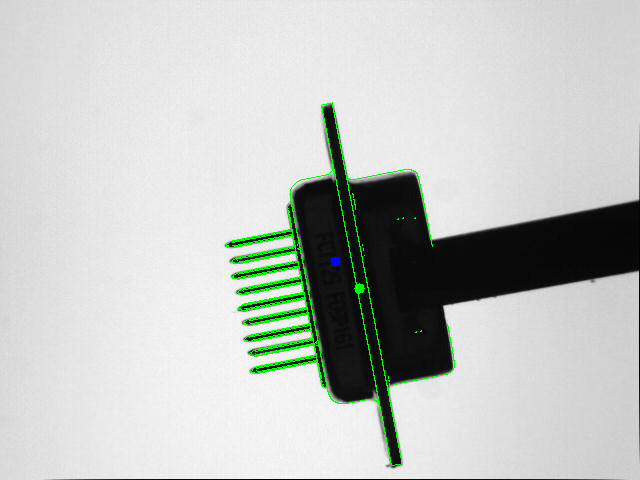}
      \caption{ -10 degree angle and 2.5mm offset in X and Y.}
      \label{fig:poseestimation:2}
    \end{subfigure}%

    \begin{subfigure}[t]{.20\textwidth}
      \centering
      \includegraphics[trim=50 0 50 0,clip,width=0.99\linewidth]{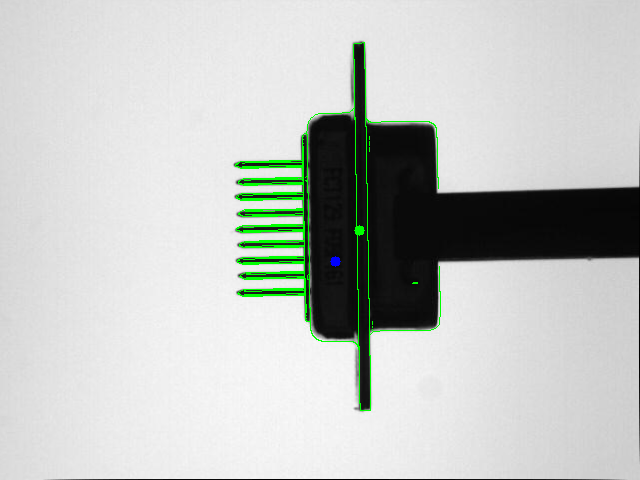}
      \caption{ 0 degree angle and 2.5mm offset in X and Y. }
      \label{fig:poseestimation:3}
    \end{subfigure}%
    ~
    \begin{subfigure}[t]{.20\textwidth}
      \centering
      \includegraphics[trim=50 0 50 0,clip,width=0.99\linewidth]{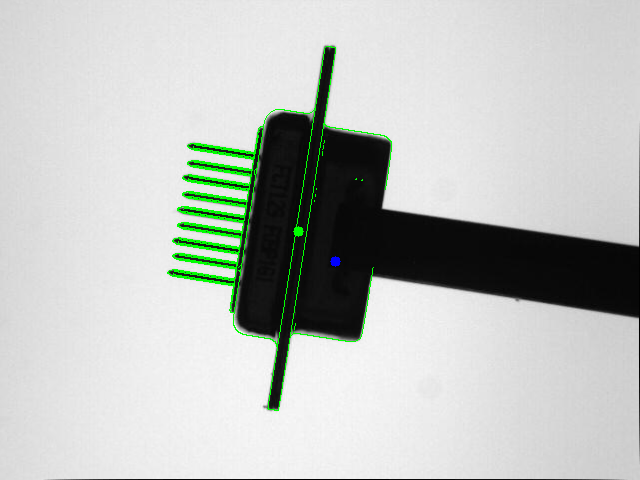}
      \caption{10 degree angle and 2.5mm offset in X and Y.}
      \label{fig:poseestimation:4}
    \end{subfigure}%
    
    \caption{Pose estimation performed on the test dataset.}
    \label{fig:poseestimation}
    \end{center}
\end{figure}

By comparing $\ _{cam} T ^{obj-expected}$ with a new pose estimate at the position, the offset in position can be found. An illustration of the different poses and the resulting pose estimation is shown in Fig.~\ref{fig:poseestimation}. The resulting offset between the expected pose and the pose estimations is shown in Tab.~\ref{tab:pose_estimation}. For all objects, the errors is at all times less than one millimetre.
%

\begin{table}[tb]
\begin{center}
    \caption{ Error for pose estimation }
    \label{tab:pose_estimation}
    \small
\begin{tabular}{|l|c|c|}
    \hline
        Object & Mean (mm) & Max (mm) \\ \hline
        \textbf{LED} & 0.326 & 0.575 \\ \hline
        \textbf{Pinheader} & 0.260 & 0.456 \\ \hline
        \textbf{D-Sub} & 0.267 & 0.529 \\ \hline
    \end{tabular}
\end{center}
\vspace{-0mm}
\end{table}


\subsection{Pin Inspection}

\begin{figure}[tb]
    \vspace{1.5mm}
    \begin{center}
    \begin{subfigure}[t]{.20\textwidth}
      \centering
      \includegraphics[trim=50 0 50 0,clip,width=0.99\linewidth]{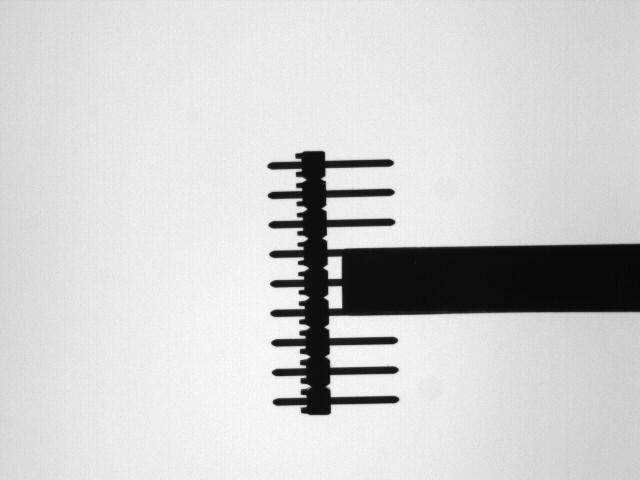}
      \caption{Straight pins.}
      \label{fig:pin_check_image:1}
    \end{subfigure}%
    ~
    \begin{subfigure}[t]{.20\textwidth}
      \centering
      \includegraphics[trim=50 0 50 0,clip,width=0.99\linewidth]{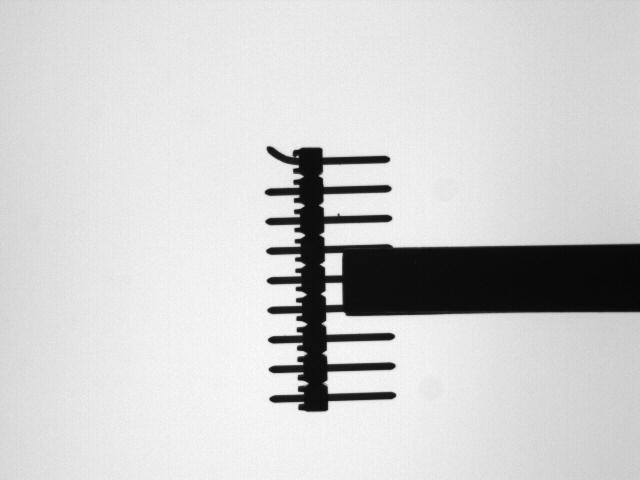}
      \caption{Bent pin.}
      \label{fig:pin_check_image:2}
    \end{subfigure}%
    
    \caption{ Visualization of straight and bent pins for the \textbf{Pinheader}. }
    \label{fig:pin_check_image}
    \end{center}
\end{figure}

To test the pin inspection a dataset of components with bent and straight pins has been generated. Different pins have been bent to a degree that the object cannot be inserted. An example of images from the dataset is shown in Fig.~\ref{fig:pin_check_image}.


\begin{figure}[tb]
    \vspace{1.5mm}
    \begin{center}
    \includegraphics[width=0.95\linewidth]{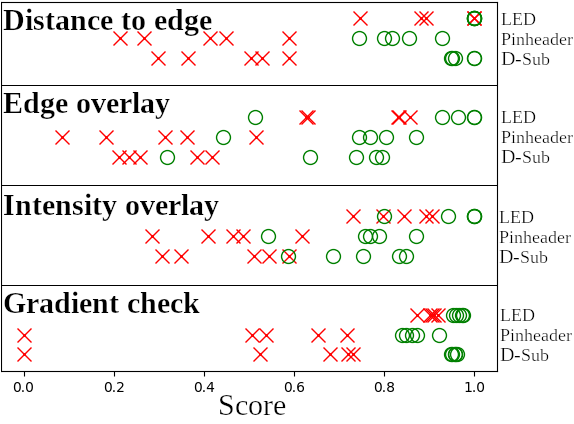} 
    \caption{ The different pin inspection scores for the three objects. Bent pins are shown as red crosses and straight pins are shown as green circles. }
    \label{fig:pin_check}
    \end{center}
\end{figure}

The resulting score for each of the different pin inspection methods is shown in Fig.~\ref{fig:pin_check}. From the scores, the two methods \textit{intensity overlay} and \textit{edge overlay} do not separate bent and straight pins well. The \textit{distance to edge} works with \textbf{Pinheader} and \textbf{D-Sub}, but fails with \textbf{LED}. The \textit{gradient check} method is able to separate bent and straight pins, but the cutoff value is not similar for the different objects. 

This could stem from inaccuracies in the CAD models. A custom cutoff value is currently necessary for each object.

\subsection{Complete system test}

To test the full usability of the system an insertion test is performed. For each of the three components, ten objects are tested. Five of the objects have bent pins and should be rejected by the system. The remaining five should be inserted successfully. 

For all objects with straight pins, the insertion was successful. The bent components were all rejected by the pin inspection. With the maximum value for a bent pin and minimum value for a straight pin at \textbf{LED:} 0.82/0.96, \textbf{Pinheader:} 0.49/0.82 and \textbf{D-Sub:} 0.0/0.87.



To further illustrate the system a visual correction is also performed. By computing $_{tcp}T^{obj-current}$ as in Eq.~\ref{eqn:tcp2obj}, the robot pose to  the object can be placed in the position from the recorded insertion pose, as in Eq.~\ref{eqn:camera}.

\begin{equation}
\label{eqn:camera}
    \ _{base}T^{tcp-comp} = \ _{base}T^{obj} (\ _{tcp}T^{obj-current})^{-1}  
\end{equation}

Furthermore, as the pin-hole sizes are known the pin-holes can be projected into the image using Eq.~\ref{eqn:projection}. It can then be visually verified that the pose correction and pin inspection will allow for the insertion of the object.


\begin{equation}
\label{eqn:projection}
    x = X \frac{f}{Z}  
\end{equation}

\begin{figure}[tb]
    \vspace{1.5mm}
    \begin{center}
    \begin{subfigure}[t]{.20\textwidth}
      \centering
      \includegraphics[trim=50 0 50 0,clip,width=0.99\linewidth]{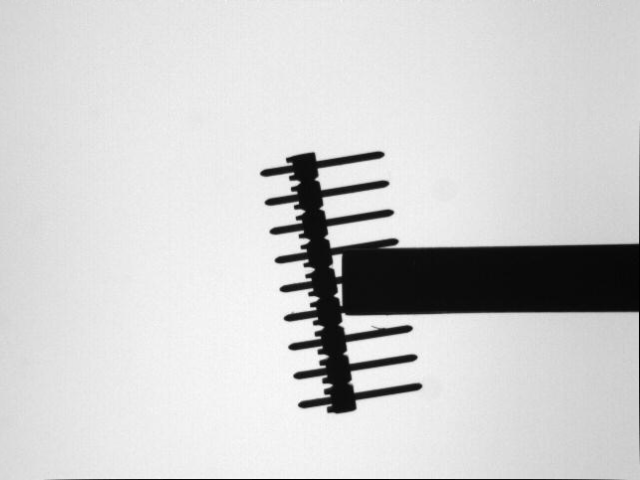}
      \caption{Initial Grasp.}
      \label{fig:insertion:1}
    \end{subfigure}%
    ~
    \begin{subfigure}[t]{.20\textwidth}
      \centering
      \includegraphics[trim=50 0 50 0,clip,width=0.99\linewidth]{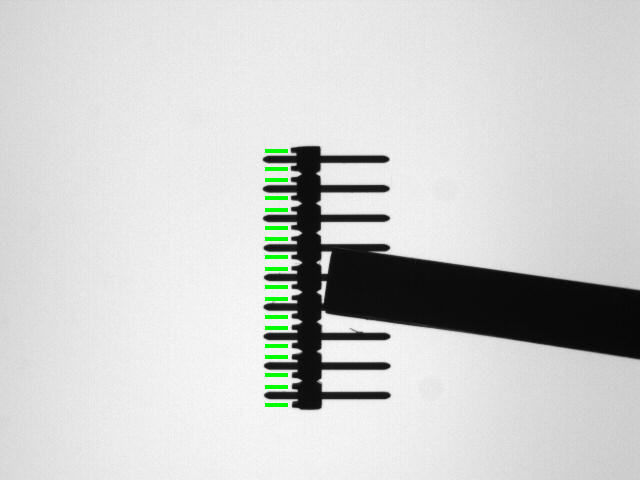}
      \caption{Corrected Pose.}
      \label{fig:insertion:2}
    \end{subfigure}%

    \begin{subfigure}[t]{.20\textwidth}
      \centering
      \includegraphics[trim=50 0 50 0,clip,width=0.99\linewidth]{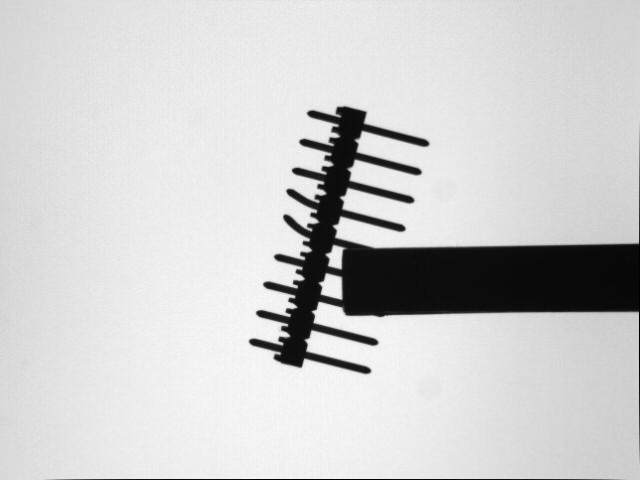}
      \caption{Initial Grasp.}
      \label{fig:insertion:3}
    \end{subfigure}%
    ~
    \begin{subfigure}[t]{.20\textwidth}
      \centering
      \includegraphics[trim=50 0 50 0,clip,width=0.99\linewidth]{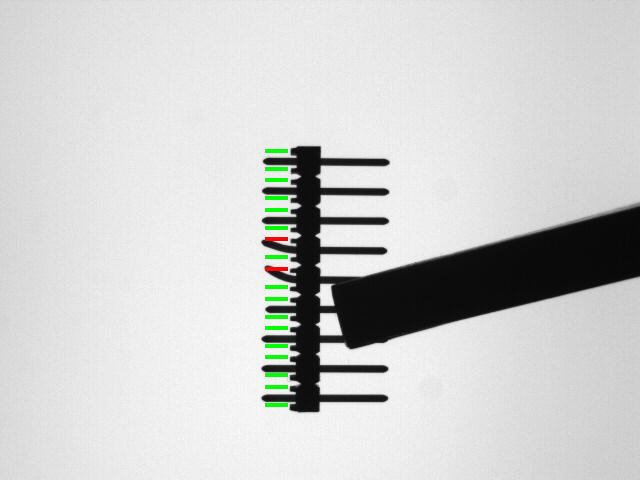}
      \caption{Corrected Pose.}
      \label{fig:insertion:4}
    \end{subfigure}%

    \begin{subfigure}[t]{.20\textwidth}
      \centering
      \includegraphics[trim=50 0 50 0,clip,width=0.99\linewidth]{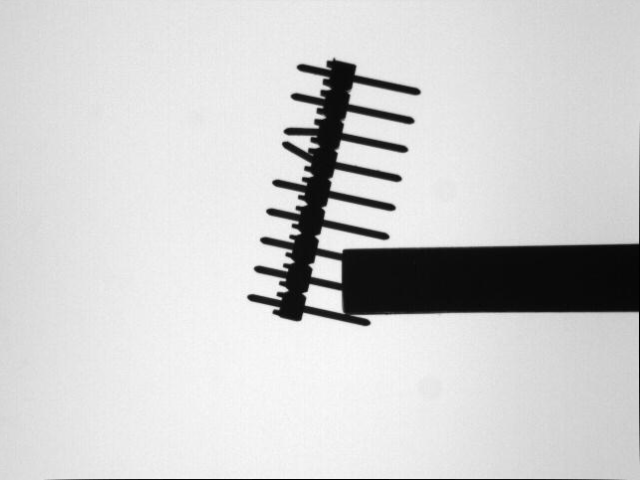}
      \caption{Initial Grasp.}
      \label{fig:insertion:5}
    \end{subfigure}%
    ~
    \begin{subfigure}[t]{.20\textwidth}
      \centering
      \includegraphics[trim=50 0 50 0,clip,width=0.99\linewidth]{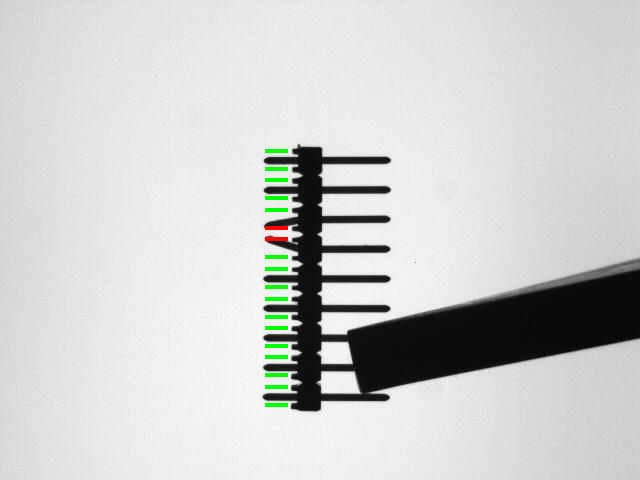}
      \caption{Corrected Pose.}
      \label{fig:insertion:6}
    \end{subfigure}%
    
    \caption{ Visualization of the pose correction and insertion for object \textbf{Pinheader}. The holes are projected into the scene based on the insertion pose and coloured green. Holes colliding with the pins are coloured red. Best viewed digitally. }
    \label{fig:insertion}
    \end{center}
\end{figure}

\begin{figure}[tb]
    \vspace{1.5mm}
    \begin{center}
    \begin{subfigure}[t]{.20\textwidth}
      \centering
      \includegraphics[trim=50 0 50 0,clip,width=0.99\linewidth]{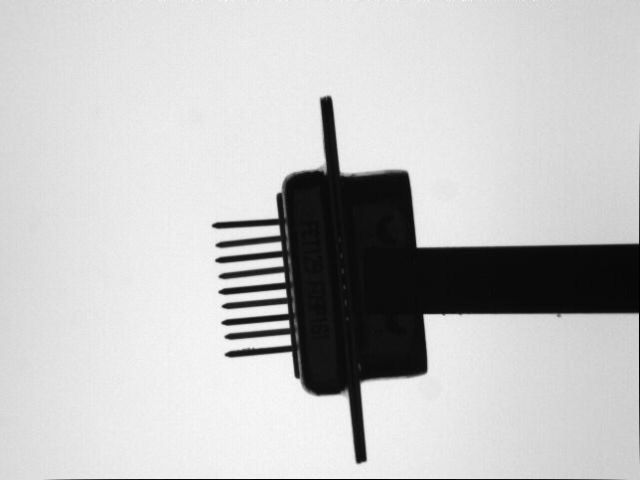}
      \caption{Initial Grasp.}
      \label{fig:insertiond:1}
    \end{subfigure}%
    ~
    \begin{subfigure}[t]{.20\textwidth}
      \centering
      \includegraphics[trim=50 0 50 0,clip,width=0.99\linewidth]{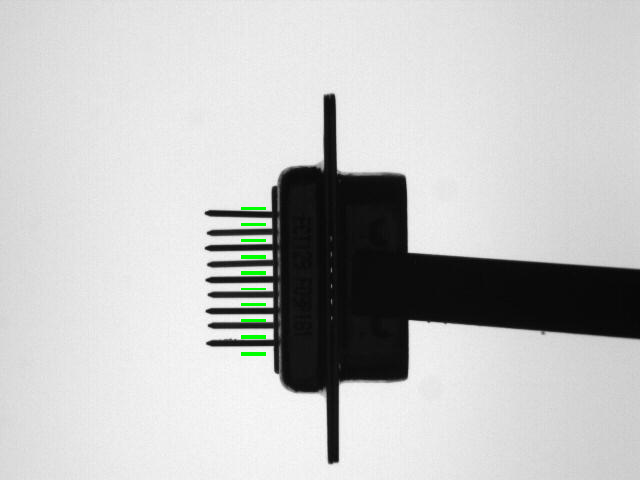}
      \caption{Corrected Pose.}
      \label{fig:insertiond:2}
    \end{subfigure}%


    \begin{subfigure}[t]{.20\textwidth}
      \centering
      \includegraphics[trim=50 0 50 0,clip,width=0.99\linewidth]{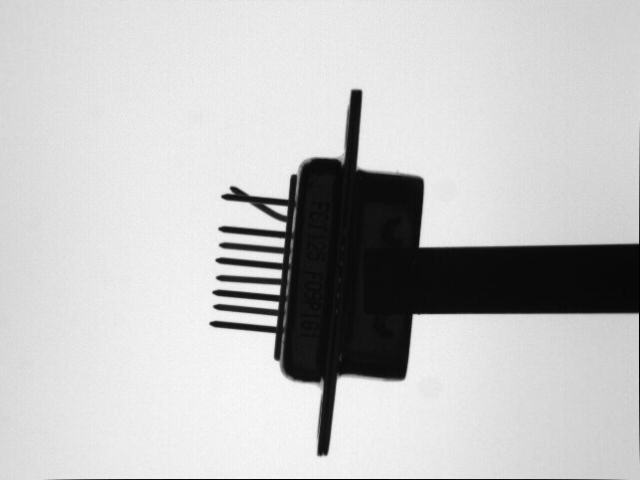}
      \caption{Initial Grasp.}
      \label{fig:insertiond:5}
    \end{subfigure}%
    ~
    \begin{subfigure}[t]{.20\textwidth}
      \centering
      \includegraphics[trim=50 0 50 0,clip,width=0.99\linewidth]{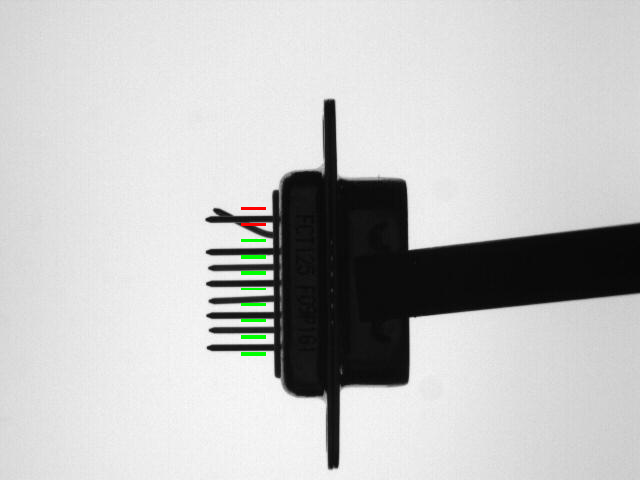}
      \caption{Corrected Pose.}
      \label{fig:insertiond:6}
    \end{subfigure}%
    
    \caption{ Visualization of the pose correction and insertion for object \textbf{D-Sub}. The holes are projected into the scene based on the insertion pose and coloured green. Holes colliding with the pins are coloured red. Best viewed digitally. }
    \label{fig:insertiond}
    \end{center}
\end{figure}

The objects are placed 0.15 meters from the camera with a focal length of 1733 pixels. As the respective hole sizes are 0.9mm, 1.2mm, 1.1mm, for the \textbf{LED}, \textbf{Pinheader}, and \textbf{D-Sub}, the resulting pixel sizes are 10, 14, and 13. In Fig.~\ref{fig:insertion} and Fig.~\ref{fig:insertiond} the visual corrections are shown the \textbf{Pinheader} and \textbf{D-Sub}, respectively.

\section{Conclusion}
\label{sec:conclusion}

In this work, we presented a novel system for in-hand pose estimation and pin inspection. The system consists of several novel contributions. A network trained for segmentation of object pins. A method to evaluate inspection poses based on robot configurations. In-hand pose estimation based on stable grasp poses. And a viable pin inspection method. Finally, the parts are connected to perform a successful insertion task.

For further work, the system could be extended with multiple cameras to inspect pins from multiple views. The pin inspection could also be performed by projecting the expected holes into the scene and checking if the object can be inserted. Furthermore, actual pose estimation of the pins could be performed, and the insertion strategy could be based on the pin positions as opposed to the object position. This would also allow for the development of more advanced insertion strategies that could potentially correct for significantly bent pins.

The in-hand pose estimation could be tested on other objects, and extended to work with multiple stable grasp poses.

\addtolength{\textheight}{-2.8in}
\newpage
\clearpage
\newpage
\clearpage

\section*{Acknowledgement}
This project was funded in part by Innovation Fund Denmark through the project MADE FAST, in part by the SDU I4.0-Lab.

\bibliographystyle{IEEEtran}
\bibliography{IEEEabrv,references}

\end{document}